\documentclass{llncs}
\usepackage{amsmath,graphicx}
\usepackage{epsfig}
\usepackage{graphicx}
\usepackage{amsmath}
\usepackage{amssymb}
\usepackage{algorithm}
\usepackage[noend]{algpseudocode}
\usepackage[utf8]{inputenc}
\usepackage[T1]{fontenc}

  \makeatletter
  \let\MYcaption\@makecaption
  \makeatother
  \usepackage{subcaption}  

\usepackage[title]{appendix}

\makeatletter
\renewcommand{\ALG@name}{{Alg.}}
\makeatother

\usepackage{tikz,pgfplots}
\usetikzlibrary{plotmarks,shapes,arrows}
\pgfplotsset{compat=newest} 
\pgfkeys{/pgf/number format/.cd,1000 sep={}}

\newlength\figureheight
\newlength\figurewidth

\def\middot{\textperiodcentered\ }

\newcommand{\fromto}[2]{$#1\!\!\rightarrow\!\!#2$}
\newcommand{\slfrac}[2]{\left.#1\middle/#2\right.}

  \newcommand{\eg}{\textit{e.g.}}
  \newcommand{\ie}{\textit{i.e.}}
  
  \newcommand{\etc}{\textit{etc.}}
  \newcommand{\etal}{\textit{et~al.}}

  \usepackage{xcolor}

  \graphicspath{{./fig1/}{./img/}}

  \usepackage{hyperref}
  \hypersetup{
    bookmarks=true,
    pdfstartview={FitH},
    colorlinks=true,
    linkcolor=black,
    citecolor=black,
    filecolor=black,
    urlcolor=black
  }

  \makeatletter
  \g@addto@macro{\UrlBreaks}{\UrlOrds}
  \makeatother

  \newlength{\boxwidth}

\title{Recursive Chaining of Reversible Image-to-Image Translators for Face Aging }
       
\author{Ari Heljakka{\inst{1,2}} \and Arno Solin\inst{2} \and Juho Kannala\inst{2}}       
\institute{GenMind Ltd, Finland \and Department of Computer Science, Aalto University, Espoo, Finland \\ \email{\{ari.heljakka,arno.solin,juho.kannala\}aalto.fi}}

\begin{document}

\maketitle

\begin{abstract}
This paper addresses the modeling and simulation of progressive changes over time, such as human face aging. By treating the age phases as a sequence of image domains, we construct a chain of transformers that map images from one age domain to the next. Leveraging recent adversarial image translation methods, our approach requires no training samples of the same individual at different ages. Here, the model must be flexible enough to translate a child face to a young adult, and all the way through the adulthood to old age. We find that some transformers in the chain can be recursively applied on their own output to cover multiple phases, compressing the chain. The structure of the chain also unearths information about the underlying physical process. We demonstrate the performance of our method with precise and intuitive metrics, and visually match with the face aging state-of-the-art.
\end{abstract}
\section{Introduction}
\label{sec:intro}

Generative Adversarial Network (GAN) \cite{goodfellow2014} variants have been successful for various image generation and transformation tasks. For image-to-image translation (such as mapping sketches to photographs), they have achieved state-of-the-art results, with paired training data \cite{isola2016} and without it \cite{kim2017,zhu2017,yi2017,liu2017}.

\begin{figure}[t]
  \centering\footnotesize
  \setlength{\boxwidth}{0.159\columnwidth}

  \makebox[\boxwidth]{2--18} \hfill
  \makebox[\boxwidth]{19--29} \hfill
  \makebox[\boxwidth]{30--39} \hfill
  \makebox[\boxwidth]{40--49} \hfill
  \makebox[\boxwidth]{50--59} \hfill
  \makebox[\boxwidth]{60+} \\[3pt]

  \framebox{\includegraphics[width=\boxwidth]{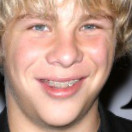}} \hfill
  \includegraphics[width=\boxwidth]{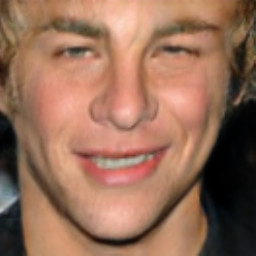} \hfill
  \includegraphics[width=\boxwidth]{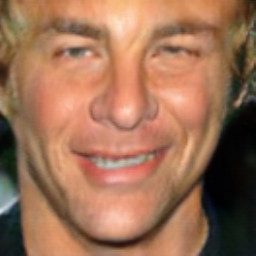} \hfill
  \includegraphics[width=\boxwidth]{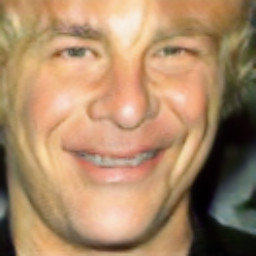} \hfill
  \includegraphics[width=\boxwidth]{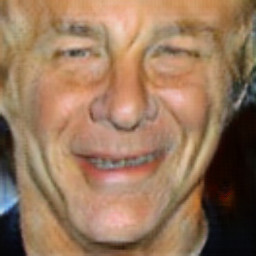} \hfill
  \includegraphics[width=\boxwidth]{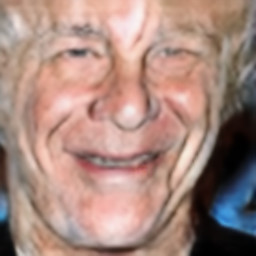} \\[1pt]
  \framebox{\includegraphics[width=\boxwidth]{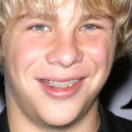}} \hfill
  \includegraphics[width=\boxwidth]{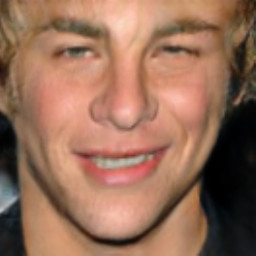} \hfill
  \includegraphics[width=\boxwidth]{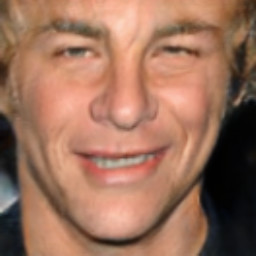} \hfill
  \includegraphics[width=\boxwidth]{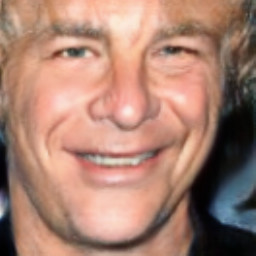} \hfill
  \includegraphics[width=\boxwidth]{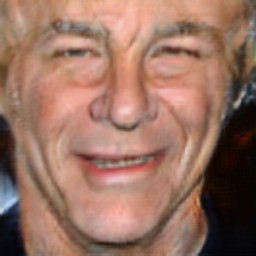} \hfill
  \includegraphics[width=\boxwidth]{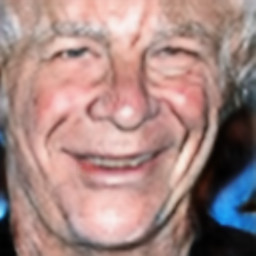} \\[1pt]
  \includegraphics[width=\boxwidth]{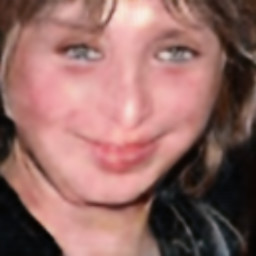} \hfill
  \includegraphics[width=\boxwidth]{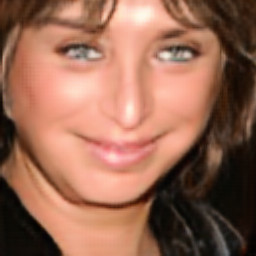} \hfill
  \includegraphics[width=\boxwidth]{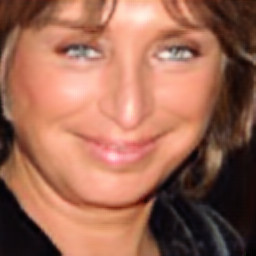} \hfill
  \includegraphics[width=\boxwidth]{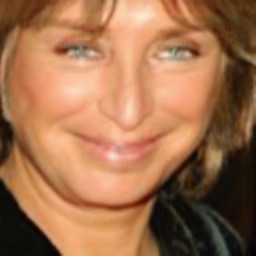} \hfill
  \framebox{\includegraphics[width=\boxwidth]{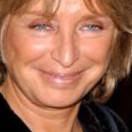}} \hfill
  \includegraphics[width=\boxwidth]{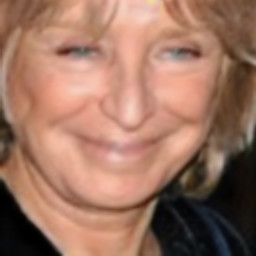} \\[1pt]
  \includegraphics[width=\boxwidth]{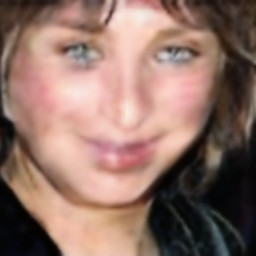} \hfill
  \includegraphics[width=\boxwidth]{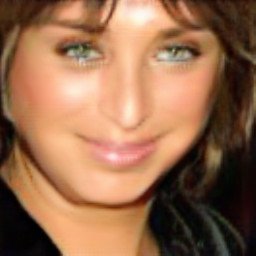} \hfill
  \includegraphics[width=\boxwidth]{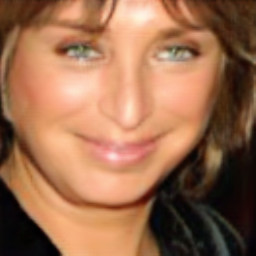} \hfill
  \includegraphics[width=\boxwidth]{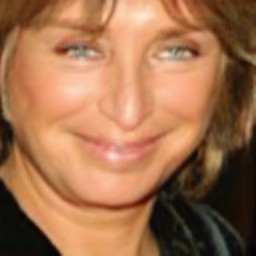} \hfill
  \framebox{\includegraphics[width=\boxwidth]{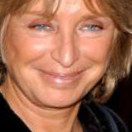}} \hfill
  \includegraphics[width=\boxwidth]{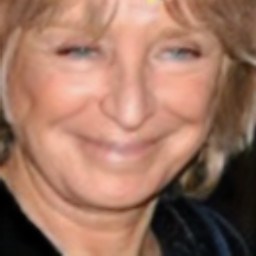} \\

  \caption{Examples of face aging transformations. Row~1: Non-recursive transformation from 15-year-old original to 65-year-old synthetic. Row~2: Partially recursive transformation by re-using the transformer of \fromto{25}{35} for also \fromto{35}{45}. Row~3: Non-recursive transform of an approximately 55-year-old original to older (65-year-old) and younger (towards 15-year-old). Row~4: Partially recursive transformation by re-using the transformer of \fromto{35}{25} for also \fromto{45}{35}.}

  \label{fig:examples}
\end{figure}

This paper generalizes image-to-image mapping to a sequential setting. Previous works have not focused on recursive application of the models on their own outputs, and there have been no extensions to apply the method for a sequence of domains, even though, \eg, \cite{choi2017} allow applying several \emph{different kinds} of domain transformations to the same image. We propose a recursive adversarial domain adaptation approach that is capable of producing step-wise transformations for human aging, as visualized in the examples in Fig.~\ref{fig:examples}. We use the \emph{reversible} image translation approach of \cite{kim2017,zhu2017,yi2017}.

Previously, Antipov \etal~\cite{antipov2017} applied a conditional GAN to simulating human face aging with good initial results. Their model distinguishes itself by explicitly enforcing the preservation of identity in face transformations. However, in reversible transformers such as ours, the preservation of identity requires no explicit measures. By learning a reversible mapping between two faces, the transformer is inclined to preserve the identity information. Face aging has been tackled before in approaches such as \cite{tiddeman2001,Kemelmacher2014,Duong2017,Lee2017}. One can also find adjustable parameters to modulate the prevalence of general attributes in an image \cite{lample2017}. While learning adjustable knobs allows for fine-grained control, our approach is suitable for robust learning of pronounced sequential changes.

The rationale of this paper follows from modeling progressive changes. Consider the development of an entity that can be approximated as a closed system (\eg, an aging human face, a deteriorating surface, a growing tree, or a changing city outline). Visual representation of such development can be approximated as a succession of snapshots taken over time, so that one snapshot is translated to the next one, then to the next one, \etc\ Such a translation requires Markovian development, so that each image carries sufficient information to enable translating it to the next stage. This allows for \eg\ face aging, but not for modeling the day--night cycle.

Utilizing unpaired image-to-image translation methods, we can learn these translation steps. Suppose we have training samples from an entity class $P$ such that we can at least roughly assign each one to a bin $P_i$ that corresponds to its developmental phase $i$. Knowing the ordering of the phases, the model can learn to transform an image from stage $P_i$ to stage $P_{i+1}$. The proposed method can simulate one such possible line of development, with the hope that the real underlying process has sufficient regularity to it. In this paper, the application of interest is human face aging, which is both a challenging modeling problem and the results are easy to validate by human readers (as humans are highly specialized in face perception).

The contributions of this paper are as follows.
\begin{itemize}
  \item We show how a chain framework of unsupervised reversible GAN transformers can be constructed to convert a human face into a desired age, along with the necessary image pre-processing steps. 

  \item We show that using an auxiliary scoring method (\eg\ a face age estimator) to rank the transformers between training stages, we can leverage transfer learning with a simple meta-training scheme that re-uses the best transformers and compresses the chain with no performance cost. The auxiliary score provides no learning gradient for an individual transformer. However, we show that the score does consistently reflect the training progress.

  \item Consequently, our ability to compress the transformer chain can be used to separate the high-level linear (\eg\ increasing formation of wrinkles) and nonlinear stages of development, producing a compact human-interpretable development descriptor.

\end{itemize}

\section{Proposed Method}
\label{sec:pmethod}

\subsection{Sequential transformations}
\label{ssec:seqdomada}

In reversible GAN transformers, such as \cite{zhu2017}, there are two generator networks and two discriminator networks. One generator transforms images from domain $A$ to domain $B$ and the other from $B$ to $A$. One discriminator separates between whether an image originates from the training set of $B$ or from the first generator. The other discriminator does the same for $A$ and the second generator. The resulting generators will have learnt to map image features between the domains.

We can treat this kind of a network as a single building block of a chain of domain transformers. For paired domains such as summer--winter or horses--zebras, the notion of such chaining does not apply. However, it does apply for classes of closed systems that evolve over time along a path of certain regularity. If we can disentangle the relevant patterns of development, a transition between two stages may occur by repeated or one-time application of those patterns.

A sequential transformation can be described as a composition of operators, with fundamental relations such as reflexivity and transitivity. Given a mapping $F$ from domain $A$ to domain $B$, what is the intended meaning of $F(F(a))$, $a \in A$? In the context of regular non-sequential transformations such as \cite{zhu2017}, a reasonable option would be to require either reflexivity as $F(F(a)) \in A$ or anti-reflexivity as $F(F(a)) \notin A$ or $F(F(a)) \in B$. These constraints could be added to existing image-to-image translators.

The more relevant case here is transitivity. For domains $A_t$, $t=\{1, 2, 3\}$, and $a_i \in A_i$, we would like to have $F(a_1) \in A_2$, and also $F(F(a_1)) \in A_3$. This allows us to enforce the transformer's ability to perform multiple transformations, turning it into an extra loss term. One may also vary the `domain thickness' with respect to $F$, \ie\ the measure of how many times F should be applied in order to move from domain $i$ to domain $i+1$ (\eg\ $2$, $1$, $1/2$, \etc). With enough domains, thickness of 1 suffices.

\subsection{Meta-training of transformer models}
\label{ssec:seqmetatr}

We train a chain of transformers to cover each developmental stage, as given in Alg.~\ref{alg:recu}. The algorithm requires a way to measure model performance at meta-level, i.e. between the actual training runs. For GANs, traditional likelihood measures are problematic \cite{theis2016,wu2016} and in some cases inversely correlated with image quality \cite{Danihelka2017}. GANs are often prone to collapse into a specific mode of the distribution, missing much of the diversity of the training data. Trainable auxiliary evaluators like \cite{Danihelka2017,lopezpaz2016} do help, but for many application domains, we already have automated deterministic scoring tools for images. For human face aging, such auxiliary evaluators exist. We chose \cite{antipov2016}.

The algorithm starts by training a separate transformer network $\Phi_1$ between the first two consequtive datasets, \eg\ 15 and 25-year-old face images, for $N$ steps ($N$ determined empirically). Transformers trained in this way are called \emph{baseline}. We use CycleGAN \cite{zhu2017}, but the algorithm is applicable for any image-to-image transformer.

Then, we train another model $\Phi_2$ in the exact same way for the next stage pair, \eg\ mapping from 25 to 35-year-olds. But now, we also re-train a copy of the model $\Phi_1$ recycled from the previous stage with additional $N/2$ steps on its earlier data and $N/2$ on the next stage data. For example, $\Phi_1$, originally trained for \fromto{15}{25} conversion, is now re-trained also on \fromto{25}{35} data.
In order to benchmark $\Phi_1$ and $\Phi_2$, we measure the perceived age of the faces in the images they produce, with the estimator \cite{antipov2016}. The transformations should age by 10 years, so we can score each transformer simply by the error in the mean age of its output distribution, normalized by its standard deviation (alg.\ lines 8--9).

If the re-trained transformer is not significantly worse on this new transformation than the baseline transformer, we discard the baseline transformer and replace it with the re-trained transformer. Otherwise we discard the re-trained one. The remaining transformer will then be tried out on the next stage, and continuing until the end of the transformer chain. By virtue of reversible transformers, we could also run the same algorithm backwards, finding the best models that make an old face young again (rows 2 and 4 in Fig.~\ref{fig:examples}).

The auxiliary scoring method is only used for a binary choice between two competing transformers during stage transitions. The score, therefore, provides no gradient for the training. Furthermore, our scoring model has been trained on an altogether different dataset and model architecture. Retrospectively, we calculate what the scores would have been during training (Fig.~\ref{fig:vperf}).

The algorithm uses a simple greedy search. By enforcing backward compatibility only to the previous transformation stage, we will only prevent forgetting in short sequences. For the six domains, we found the incurring performance penalty to be minor (Fig.~\ref{fig:vperf}). To scale up, we could simply cap the maximum number of re-uses of a model or the allowed error from forgetting, or add equal parts of even earlier domains to the training set of the re-used model.

\begin{algorithm}[!t]
  \caption{Greedy forward-mode recursive transformer chain with two-step backward-compatibility. In the experiments, we used $\mu^\mathrm{target}_i =  [15, 25, \ldots, 65]$, $N_D = 6$, $S=\text{600,000}$, $\varepsilon=0.1$.}\label{alg:recu}\footnotesize
  \begin{algorithmic}[1]
    \State {\bf Require:} Number of stages $N_D$,
    data sets $D_i$ and target mean age $\mu^\mathrm{target}_i$ with $i \in [1, N_D]$,
    trainable models $\Phi_j:\mathbb{R}^{256\times256} \to \mathbb{R}^{256\times256}$ with $j \in [1, N_D-1]$, 
    number of steps $S$, auxiliary age estimator $\Gamma: \mathbb{R}^{256\times256} \to \mathbb{R}$
  
  \State {\bf Initialize:} $a \gets 1$ \Comment{Denote the index of the $\Phi$ model we try to re-use}
  \State \phantom{\bf Initialize:} $\Phi_1 \gets \mathrm{train}(\Phi_1, [(D_1, D_2)], S)$
  \For{$i = 2,\ldots,(N_D-1)$}
   \State $\Phi_i \gets \mathrm{train}(\Phi_i, [(D_i, D_{i+1})], S)$
   \State $\Phi'_a \gets \mathrm{copy}(\Phi_a)$
     \State $\Phi'_a \gets \mathrm{train}(\Phi'_a, [(D_{i-1}, D_{i}), (D_{i}, D_{i+1})], S)$
     \State $E' \gets \left|\slfrac{\mathop{\mathbb{E}_{d \sim D_i}}[\|\Gamma(\Phi'_a(d)) - \mu^{target}_{i+1}\|_1]} {\sigma(\Gamma(\Phi'_a(d)))}\right|$
     \State $E \gets \left|\slfrac{\mathop{\mathbb{E}_{d \sim D_i}}[\|\Gamma(\Phi_i(d)) - \mu^{target}_{i+1}\|_1]} {\sigma(\Gamma(\Phi_i(d)))}\right| $
     \If{$|E - E'| < \varepsilon$} \Comment {Recycled model wins}
      \State $\mathrm{release}(\Phi_i)$
      \State $\mathrm{release}(\Phi_{a})$
      \State $\Phi_a \gets \Phi_{i} \gets \Phi'_a$ \Comment {Upgrade $\Phi_a$ and try re-using again}
     \Else
      \State $\mathrm{release}(\Phi'_a) $
      \State $a \gets i$ \Comment {Next, try re-using the most recent base model}
     \EndIf
  \EndFor

  \end{algorithmic}
\end{algorithm}

\section{Experiments}
\label{sec:apps}

\subsection{Dataset and auxiliary age estimator}
\label{ssec:dataset}
For training, we use the Cross-Age Celebrity Dataset (CACD, \cite{chen2014}), with a large number of age-annotated images. Our chosen auxiliary age estimator \cite{antipov2016}, pre-trained on the IMDB-Wiki dataset~\cite{rothe2015}, was used as-is (the estimator has not seen any of our training data).
We found it necessary to improve the CACD data in two ways. First, we used the off-the-shelf face alignment utility of \cite{schroff2015} to crop and align the faces based on landmarks. Second, despite the existing age annotations of CACD, they are not accurate enough for age-estimation \cite{cacd}. We ran the auxiliary age estimator on the data and found major discrepancy between the annotations and the results of the estimator. We also confirmed visually that the estimator was closer to the ground truth. Using the estimator as ground-truth, we re-annotated the whole CACD data accordingly. Our model, however, has access neither to the age information nor the estimator. It only knows that the pictures come from different domains, at 10-year accuracy. For validation, we use a small subset of the IMDB-Wiki dataset.

We divided the data into slots 2--18, 19--29, 30--39, 40--49, 50--59, and 60--78 (six domains, with five direct transformation paths). This enables direct comparison with \cite{antipov2017}. We removed enough samples so that the mean age in the sets is 10 years apart---15, 25, 35, 45, 55, and 65, respectively.  Since the numbers of samples may differ between domains, we measure training progress by the number of steps, not epochs, so as to maintain commensurability between stages.

\subsection{Architecture and training}
\label{ssec:mnist}
The architecture of each transformer module follows \cite{zhu2017} (which re-uses structures from \cite{johnson2016} and \cite{isola2016}), with two generators and two discriminators per module. The generators utilize convolution layers for encoding, 9 ResNet blocks for transformation, and deconvolution layers for decoding (original source code adopted from \cite{pytorch-cycle-gan}). Our image preprocessing reduces the real face resolution to $132\!\times\!132$, which is then upscaled to $256\!\times\!256$ for computational efficiency using off-the-shelf Mitchell--Netravali filter \cite{mitchell1988}. The full chain is composed of five independently trained successive modules, so that one module feeds its output to the next. The results of each transformation stage can be evaluated independently, and external input can be fed in at any stage of the chain. For recursion, one can co-train with 3 stages (\eg \fromto{15}{25} and \fromto{25}{35}), or apply the model twice (\fromto{15}{35}).

We trained with ADAM (\cite{adam}, learning rate $0.0002$, $\beta_1=0.5$, $\beta_2=0.999$) with a batch size one. As in \cite{zhu2017,shrivastava2016}, we update the discriminators using a buffer of 50 recently generated images rather than only the most recently produced ones. The training time on an NVIDIA P100 workstation was 540~h (300~h for the baseline, 60~h for each re-training session).

\subsection{Human face age progression}
\label{ssec:face}
Our solution is based on a pipeline of successive transformations. In order to train a single transformation, say \fromto{25}{35}, we train a single transformer network. By the reversible architecture, we simultaneously train the network to carry out the full inverse transformation, \fromto{35}{25}.

\begin{figure}[h!]
  \centering\footnotesize
  \pgfplotsset{yticklabel style={rotate=90},width=\figurewidth,height=\figureheight,legend style={font=\footnotesize},ytick={15,25,35,45,55,65},legend style={row sep=-3pt},grid style=dotted}
  \setlength{\figurewidth}{0.90\columnwidth}
  \setlength{\figureheight}{.6\figurewidth}
  \begin{subfigure}[t]{1.00\columnwidth}
    \centering
    \input{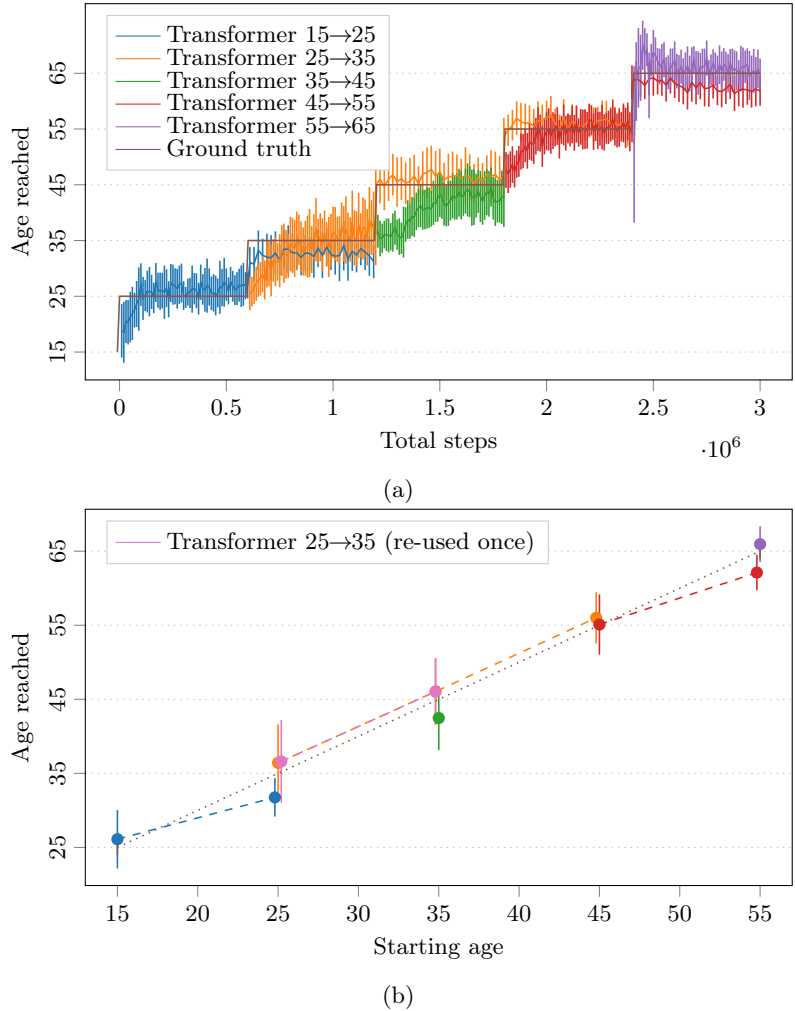}

    \caption{\ }
    \label{fig:tperf}
  \end{subfigure} 
  \begin{subfigure}[t]{1.00\columnwidth}
    \centering
    % This file was created by matplotlib2tikz v0.6.15.
\begin{tikzpicture}

\definecolor{color0}{rgb}{0.12156862745098,0.466666666666667,0.705882352941177}
\definecolor{color5}{rgb}{0.549019607843137,0.337254901960784,0.294117647058824}
\definecolor{color6}{rgb}{0.890196078431372,0.466666666666667,0.76078431372549}
\definecolor{color1}{rgb}{1,0.498039215686275,0.0549019607843137}
\definecolor{color2}{rgb}{0.172549019607843,0.627450980392157,0.172549019607843}
\definecolor{color3}{rgb}{0.83921568627451,0.152941176470588,0.156862745098039}
\definecolor{color4}{rgb}{0.580392156862745,0.403921568627451,0.741176470588235}

\begin{axis}[
xlabel={Starting age},
ylabel={Age reached},
xmin=13, xmax=57,
ymin=19.85545, ymax=70.66555,
tick align=outside,
tick pos=left,
x grid style={lightgray!92.02614379084967!black},
ymajorgrids,
y grid style={lightgray!92.02614379084967!black},
legend cell align={left},
legend style={at={(0.03,0.97)}, anchor=north west, draw=white!80.0!black},
legend entries={{Transformer \fromto{25}{35} (re-used once)}}
]
+\addlegendimage{no markers, color6}
\path [draw=color0, semithick] (axis cs:15,22.165)
--(axis cs:15,30.055);

\path [draw=color0, semithick] (axis cs:24.8,29.167)
--(axis cs:24.8,34.353);

\path [draw=color1, semithick] (axis cs:25,31.235)
--(axis cs:25,41.605);

\path [draw=color1, semithick] (axis cs:34.8,41.543)
--(axis cs:34.8,50.557);

\path [draw=color1, semithick] (axis cs:44.8,52.538)
--(axis cs:44.8,59.462);

\path [draw=color2, semithick] (axis cs:35,38.137)
--(axis cs:35,46.803);

\path [draw=color3, semithick] (axis cs:45,51.025)
--(axis cs:45,59.175);

\path [draw=color3, semithick] (axis cs:54.8,59.707)
--(axis cs:54.8,64.513);

\path [draw=color4, semithick] (axis cs:55,63.544)
--(axis cs:55,68.356);

\path [draw=color5, semithick] (axis cs:15,24)
--(axis cs:15,26);

\path [draw=color5, semithick] (axis cs:55,64)
--(axis cs:55,66);

\path [draw=color6, semithick] (axis cs:25.2,31.039)
--(axis cs:25.2,42.181);

\path [draw=color6, semithick] (axis cs:34.8,41.543)
--(axis cs:34.8,50.557);

\addplot [semithick, color0, dashed, mark=*, mark size=2, mark options={solid}, forget plot]
table {%
15 26.11
24.8 31.76
};
\addplot [semithick, color1, dashed, mark=*, mark size=2, mark options={solid}, forget plot]
table {%
25 36.42
34.8 46.05
44.8 56
};
\addplot [semithick, color2, mark=*, mark size=2, mark options={solid}, only marks, forget plot]
table {%
35 42.47
};
\addplot [semithick, color3, dashed, mark=*, mark size=2, mark options={solid}, forget plot]
table {%
45 55.1
54.8 62.11
};
\addplot [semithick, color4, mark=*, mark size=2, mark options={solid}, only marks, forget plot]
table {%
55 65.95
};
\addplot [semithick, color5, dotted, mark=asterisk*, mark size=2, mark options={solid}, forget plot]
table {%
15 25
55 65
};
\addplot [semithick, color6, dashed, mark=*, mark size=2, mark options={solid}, forget plot]
table {%
25.2 36.61
34.8 46.05
};
\end{axis}

\end{tikzpicture}
    \caption{\ }   
    \label{fig:vperf}    
  \end{subfigure}  
  \caption{{\bf (a)}~Performance of modules as a function of training steps (with stage-specific transformation target), greedily selecting the best-performing models for re-use. The error bars indicate the standard deviation of the ages of generated test samples. Some of the variance is by design, as we intentionally challenge the model with a wide \emph{range} of ages for robustness and generalization in training data (for clarity of visualization, the ground truth variance not shown).
  {\bf (b)}~Resulting age after transformation in validation set, as a function of the input age. During training, the model learned that \fromto{25}{35} transformer can be recycled for \fromto{35}{45} to transform better than the freshly trained \fromto{35}{45} transformer. The \fromto{35}{45} transformer was thus discarded (see Alg.~\ref{alg:recu}). \fromto{15}{25} and \fromto{45}{55} transformers succeeded on their baseline but were not re-usable.}
\end{figure}

We trained and evaluated each transformer network according to Alg.~\ref{alg:recu}. For human face aging, our hypothesis was that the network trained for transforming faces from \fromto{15}{25} would not generalize to the other stages, whereas the network that transforms from \fromto{25}{35} would generalize to multiple later stages as well. We confirm this both during training-time (Fig.~\ref{fig:tperf}) and validation (Fig.~\ref{fig:vperf}).  The algorithm drops the baseline transformer \fromto{35}{45} off the chain ($\epsilon > 0.35$ would drop the next one, too). From this result, we can read off the development descriptor of the form $F(\text{age}=65)=F_\text{\fromto{55}{65}}(F_\text{\fromto{45}{55}}(F_\text{\fromto{25}{35}}^2(F_\text{\fromto{15}{25}}(\text{age}=15))))$. We expected that if we can re-use some transformers, this provides us with high-level information about the underflying development pattern. While our result alone only gives tentative support to this idea, we expect this approach to be also applicable with more fine-grained precision.
\vspace{-1mm}

\subsection{Comparison}
\label{ssec:comparison}
\begin{figure*}[p]
  \centering\footnotesize

  \setlength{\boxwidth}{0.13\columnwidth}

  \makebox[\boxwidth]{2--18} \hspace*{3pt}
  \makebox[\boxwidth]{19--29} \hspace*{3pt}
  \makebox[\boxwidth]{30--39} \hspace*{3pt}
  \makebox[\boxwidth]{40--49} \hspace*{3pt}
  \makebox[\boxwidth]{50--59} \hspace*{3pt}
  \makebox[\boxwidth]{60+} \\[3pt]

  \framebox{\includegraphics[width=\boxwidth]{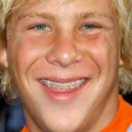}} \hspace*{3pt}
  \includegraphics[width=\boxwidth]{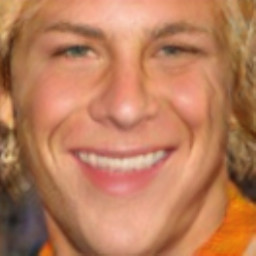} \hspace*{3pt}
  \includegraphics[width=\boxwidth]{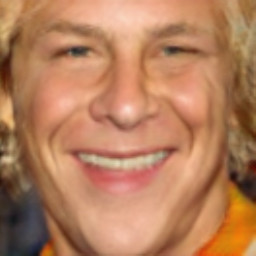} \hspace*{3pt}
  \includegraphics[width=\boxwidth]{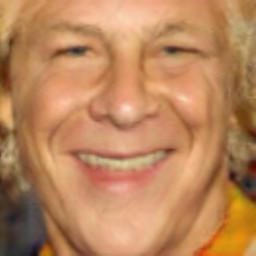} \hspace*{3pt}
  \includegraphics[width=\boxwidth]{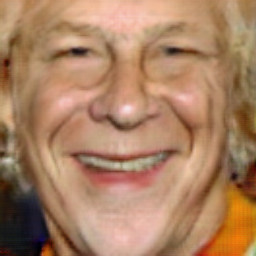} \hspace*{3pt}
  \includegraphics[width=\boxwidth]{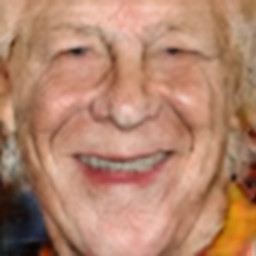} \\[1pt]
  \framebox{\includegraphics[width=\boxwidth]{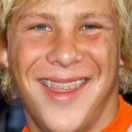}} \hspace*{3pt}
  \includegraphics[width=\boxwidth]{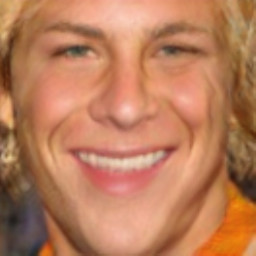} \hspace*{3pt}
  \includegraphics[width=\boxwidth]{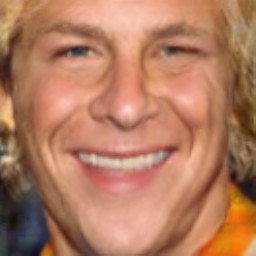} \hspace*{3pt}
  \includegraphics[width=\boxwidth]{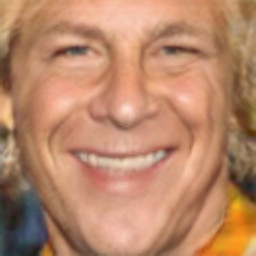} \hspace*{3pt}
  \includegraphics[width=\boxwidth]{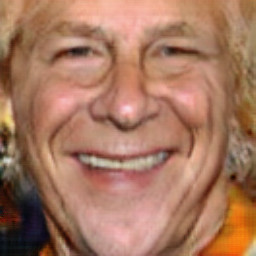} \hspace*{3pt}
  \includegraphics[width=\boxwidth]{baseline/age_65_id_3} \\[1pt]
  \framebox{\includegraphics[width=\boxwidth]{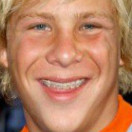}} \hspace*{3pt}
  \includegraphics[width=\boxwidth]{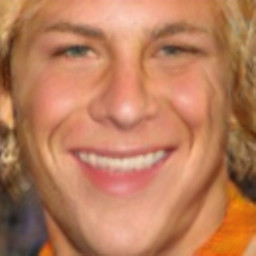} \hspace*{3pt}
  \includegraphics[width=\boxwidth]{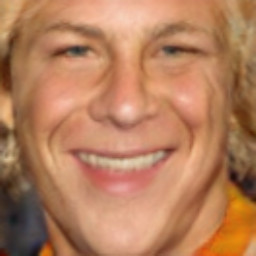} \hspace*{3pt}
  \includegraphics[width=\boxwidth]{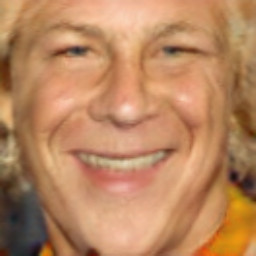} \hspace*{3pt}
  \includegraphics[width=\boxwidth]{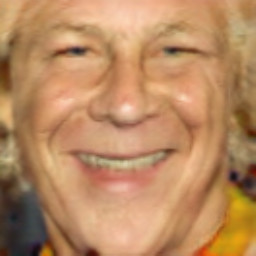} \hspace*{3pt}
  \includegraphics[width=\boxwidth]{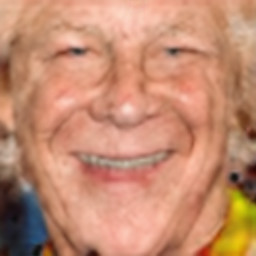} \\[10pt]
  \framebox{\includegraphics[width=\boxwidth]{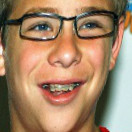}} \hspace*{3pt}
  \includegraphics[width=\boxwidth]{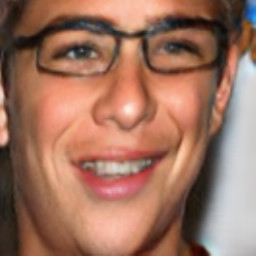} \hspace*{3pt}
  \includegraphics[width=\boxwidth]{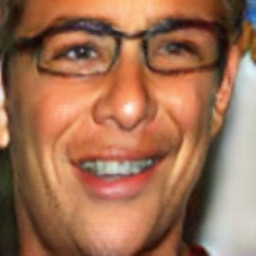} \hspace*{3pt}
  \includegraphics[width=\boxwidth]{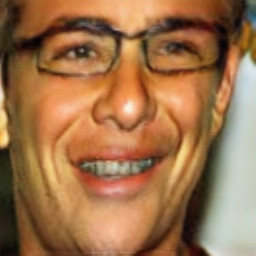} \hspace*{3pt}
  \includegraphics[width=\boxwidth]{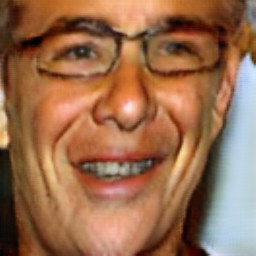} \hspace*{3pt}
  \includegraphics[width=\boxwidth]{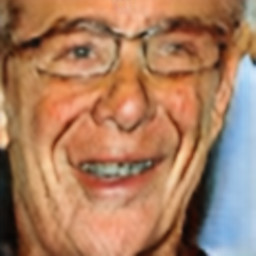} \\[1pt]
  \framebox{\includegraphics[width=\boxwidth]{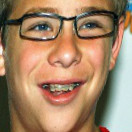}} \hspace*{3pt}
  \includegraphics[width=\boxwidth]{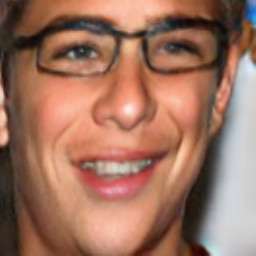} \hspace*{3pt}
  \includegraphics[width=\boxwidth]{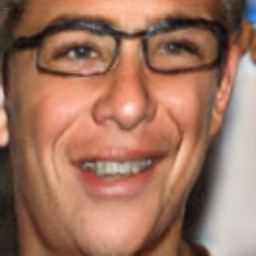} \hspace*{3pt}
  \includegraphics[width=\boxwidth]{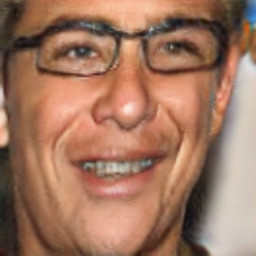} \hspace*{3pt}
  \includegraphics[width=\boxwidth]{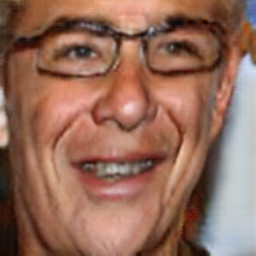} \hspace*{3pt}
  \includegraphics[width=\boxwidth]{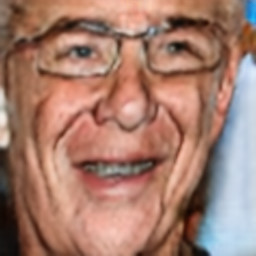} \\[1pt]
  \framebox{\includegraphics[width=\boxwidth]{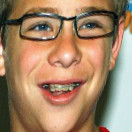}} \hspace*{3pt}
  \includegraphics[width=\boxwidth]{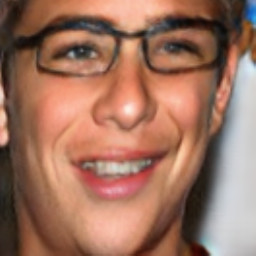} \hspace*{3pt}
  \includegraphics[width=\boxwidth]{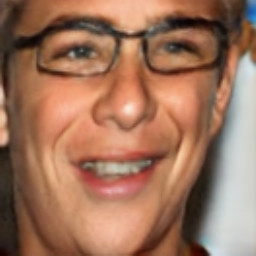} \hspace*{3pt}
  \includegraphics[width=\boxwidth]{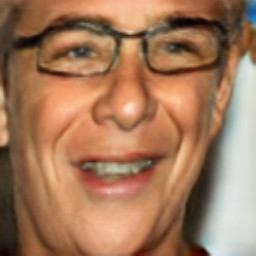} \hspace*{3pt}
  \includegraphics[width=\boxwidth]{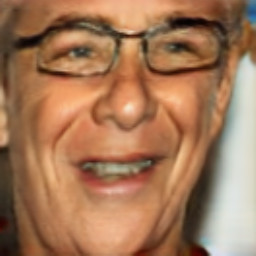} \hspace*{3pt}
  \includegraphics[width=\boxwidth]{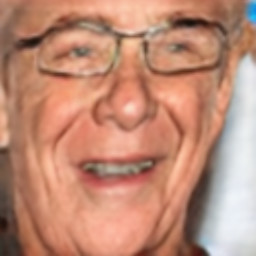} \\[10pt]
  \framebox{\includegraphics[width=\boxwidth]{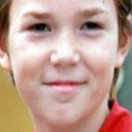}} \hspace*{3pt}
  \includegraphics[width=\boxwidth]{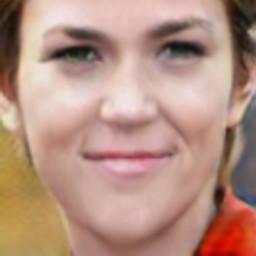} \hspace*{3pt}
  \includegraphics[width=\boxwidth]{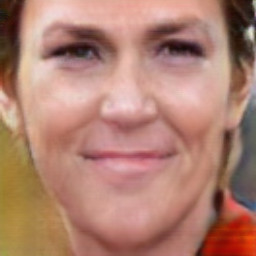} \hspace*{3pt}
  \includegraphics[width=\boxwidth]{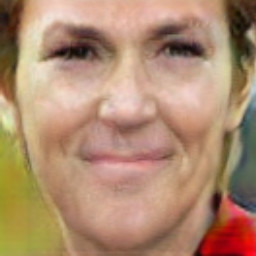} \hspace*{3pt}
  \includegraphics[width=\boxwidth]{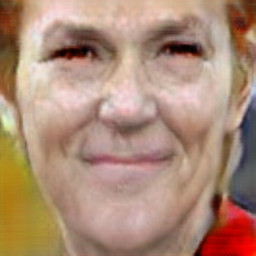} \hspace*{3pt}
  \includegraphics[width=\boxwidth]{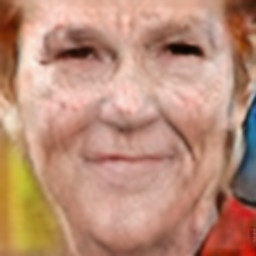} \\[1pt]
  \framebox{\includegraphics[width=\boxwidth]{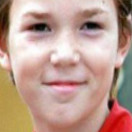}} \hspace*{3pt}
  \includegraphics[width=\boxwidth]{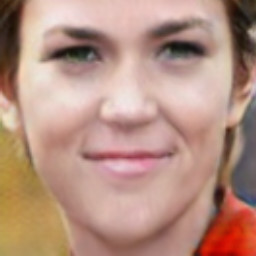} \hspace*{3pt}
  \includegraphics[width=\boxwidth]{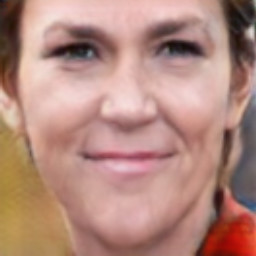} \hspace*{3pt}
  \includegraphics[width=\boxwidth]{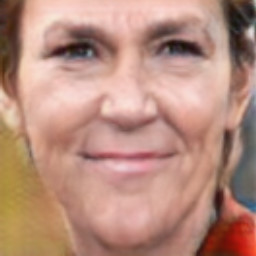} \hspace*{3pt}
  \includegraphics[width=\boxwidth]{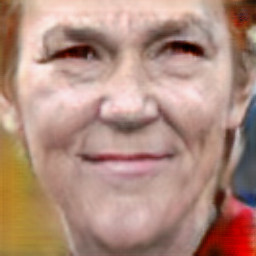} \hspace*{3pt}
  \includegraphics[width=\boxwidth]{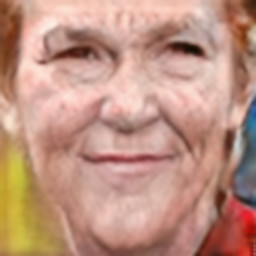} \\[1pt]
  \framebox{\includegraphics[width=\boxwidth]{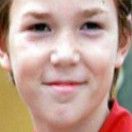}} \hspace*{3pt}
  \includegraphics[width=\boxwidth]{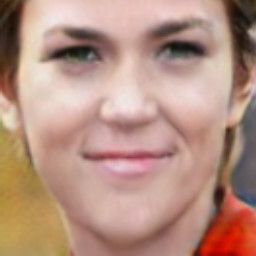} \hspace*{3pt}
  \includegraphics[width=\boxwidth]{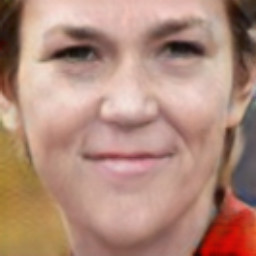} \hspace*{3pt}
  \includegraphics[width=\boxwidth]{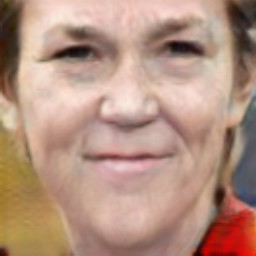} \hspace*{3pt}
  \includegraphics[width=\boxwidth]{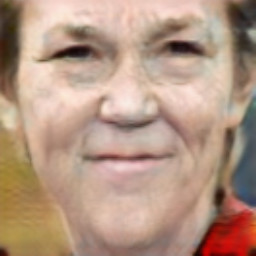} \hspace*{3pt}
  \includegraphics[width=\boxwidth]{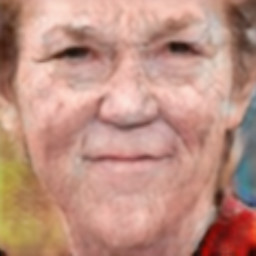} \\[1pt]
  \caption{Examples of transforming an approximately 15-year-old to 65-year-old. For each identity, row~1 shows the non-recursive transformation (applying the baseline transformer on each stage). Row~2 shows the partially-recursive transformation, with the double-trained \fromto{25}{35} transformer applied also to \fromto{35}{45} (the best chain, according to Alg.~\ref{alg:recu}). Row~3 shows the transformation with the most recursive steps, with the triple-trained \fromto{25}{35} transformer applied also to both \fromto{35}{45} and \fromto{45}{55} (the most compact chain, picked manually).}
  \label{fig:examples2}
\end{figure*}

\begin{figure*}[p]
  \centering\footnotesize
  \setlength{\boxwidth}{0.13\columnwidth}

  \makebox[\boxwidth]{2--18} \hspace*{3pt}
  \makebox[\boxwidth]{19--29} \hspace*{3pt}
  \makebox[\boxwidth]{30--39} \hspace*{3pt}
  \makebox[\boxwidth]{40--49} \hspace*{3pt}
  \makebox[\boxwidth]{50--59} \hspace*{3pt}
  \makebox[\boxwidth]{60+} \\[3pt]

  \includegraphics[width=\boxwidth]{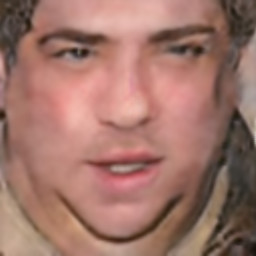} \hspace*{3pt}
  \includegraphics[width=\boxwidth]{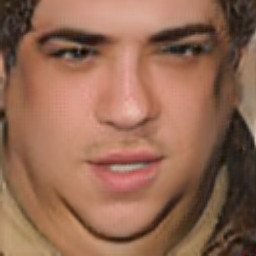} \hspace*{3pt}
  \includegraphics[width=\boxwidth]{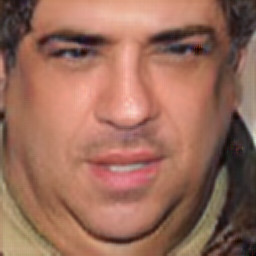} \hspace*{3pt}
  \includegraphics[width=\boxwidth]{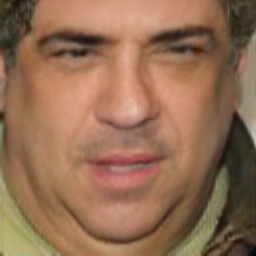} \hspace*{3pt}
  \framebox{\includegraphics[width=\boxwidth]{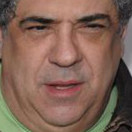}} \hspace*{3pt}
  \includegraphics[width=\boxwidth]{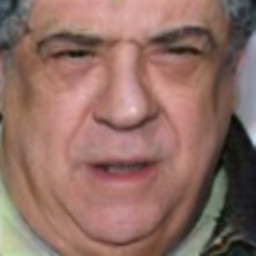} \\[1pt]

  \includegraphics[width=\boxwidth]{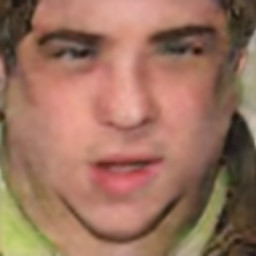} \hspace*{3pt}
  \includegraphics[width=\boxwidth]{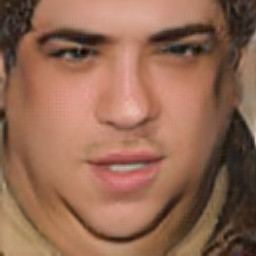} \hspace*{3pt}
  \includegraphics[width=\boxwidth]{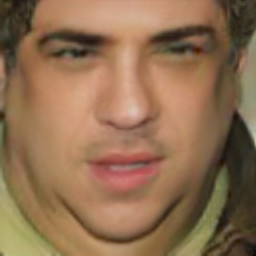} \hspace*{3pt}
  \includegraphics[width=\boxwidth]{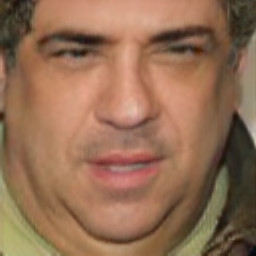} \hspace*{3pt}
  \framebox{\includegraphics[width=\boxwidth]{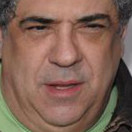}} \hspace*{3pt}
  \includegraphics[width=\boxwidth]{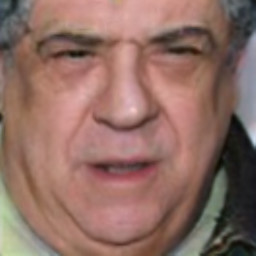} \\[1pt]

  \includegraphics[width=\boxwidth]{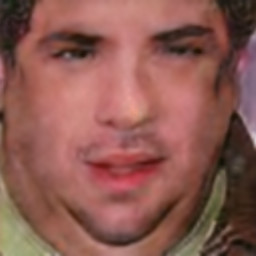} \hspace*{3pt}
  \includegraphics[width=\boxwidth]{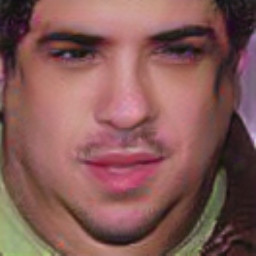} \hspace*{3pt}
  \includegraphics[width=\boxwidth]{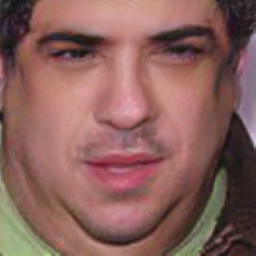} \hspace*{3pt}
  \includegraphics[width=\boxwidth]{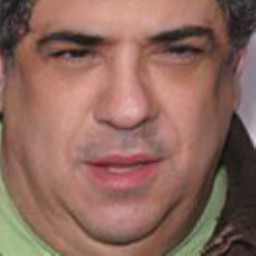} \hspace*{3pt}
  \framebox{\includegraphics[width=\boxwidth]{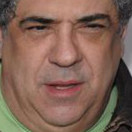}} \hspace*{3pt}
  \includegraphics[width=\boxwidth]{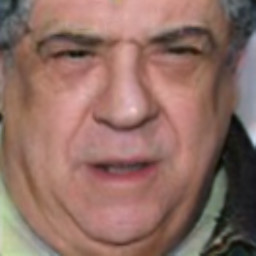} \\[10pt]

  \includegraphics[width=\boxwidth]{baseline/from_55/age_15_id_8} \hspace*{3pt}
  \includegraphics[width=\boxwidth]{baseline/from_55/age_25_id_8} \hspace*{3pt}
  \includegraphics[width=\boxwidth]{baseline/from_55/age_35_id_8} \hspace*{3pt}
  \includegraphics[width=\boxwidth]{baseline/from_55/age_45_id_8} \hspace*{3pt}
  \framebox{\includegraphics[width=\boxwidth]{baseline/from_55/age_55_id_8}} \hspace*{3pt}
  \includegraphics[width=\boxwidth]{baseline/from_55/age_65_id_8} \\[1pt]
  \includegraphics[width=\boxwidth]{best/from_55/age_15_id_8} \hspace*{3pt}
  \includegraphics[width=\boxwidth]{best/from_55/age_25_id_8} \hspace*{3pt}
  \includegraphics[width=\boxwidth]{best/from_55/age_35_id_8} \hspace*{3pt}
  \includegraphics[width=\boxwidth]{best/from_55/age_45_id_8} \hspace*{3pt}
  \framebox{\includegraphics[width=\boxwidth]{best/from_55/age_55_id_8}} \hspace*{3pt}
  \includegraphics[width=\boxwidth]{best/from_55/age_65_id_8} \\[1pt]
  \includegraphics[width=\boxwidth]{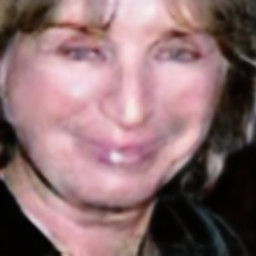} \hspace*{3pt}
  \includegraphics[width=\boxwidth]{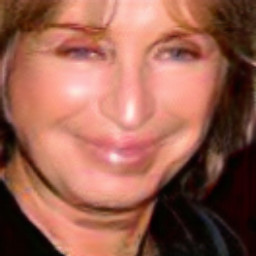} \hspace*{3pt}
  \includegraphics[width=\boxwidth]{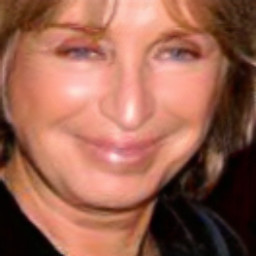} \hspace*{3pt}
  \includegraphics[width=\boxwidth]{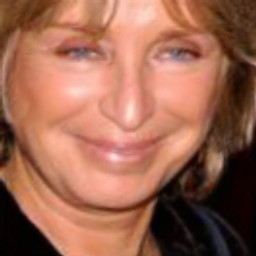} \hspace*{3pt}
  \framebox{\includegraphics[width=\boxwidth]{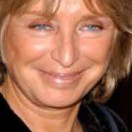}} \hspace*{3pt}
  \includegraphics[width=\boxwidth]{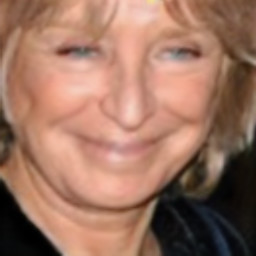} \\[10pt]

  \includegraphics[width=\boxwidth]{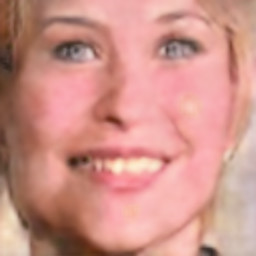} \hspace*{3pt}
  \includegraphics[width=\boxwidth]{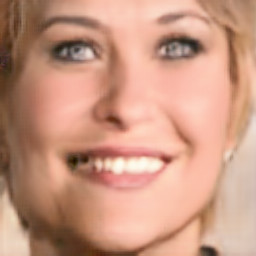} \hspace*{3pt}
  \includegraphics[width=\boxwidth]{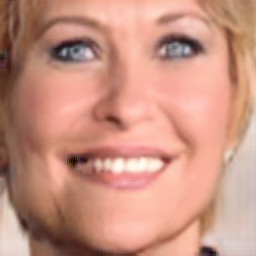} \hspace*{3pt}
  \includegraphics[width=\boxwidth]{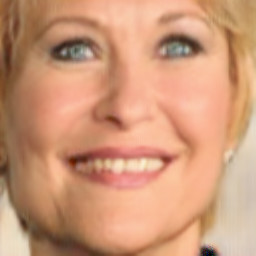} \hspace*{3pt}
  \framebox{\includegraphics[width=\boxwidth]{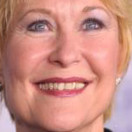}} \hspace*{3pt}
  \includegraphics[width=\boxwidth]{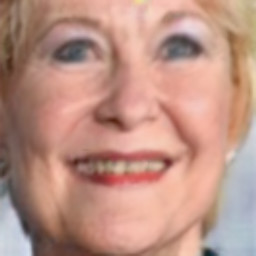} \\[1pt]
  \includegraphics[width=\boxwidth]{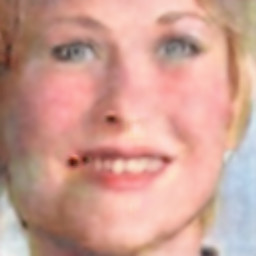} \hspace*{3pt}
  \includegraphics[width=\boxwidth]{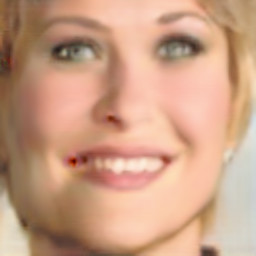} \hspace*{3pt}
  \includegraphics[width=\boxwidth]{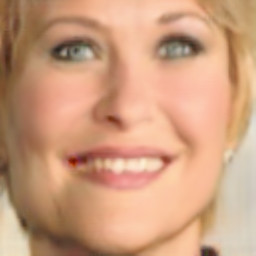} \hspace*{3pt}
  \includegraphics[width=\boxwidth]{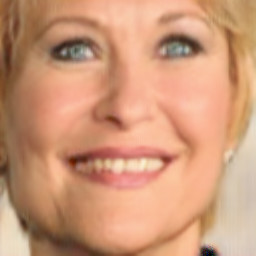} \hspace*{3pt}
  \framebox{\includegraphics[width=\boxwidth]{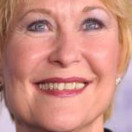}} \hspace*{3pt}
  \includegraphics[width=\boxwidth]{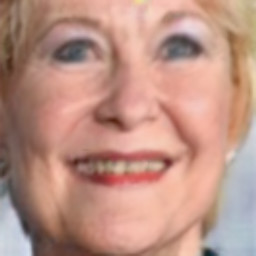} \\[1pt]
  \includegraphics[width=\boxwidth]{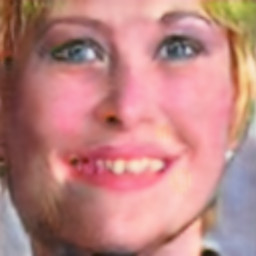} \hspace*{3pt}
  \includegraphics[width=\boxwidth]{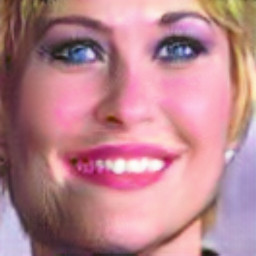} \hspace*{3pt}
  \includegraphics[width=\boxwidth]{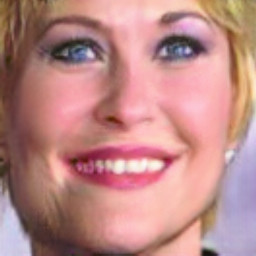} \hspace*{3pt}
  \includegraphics[width=\boxwidth]{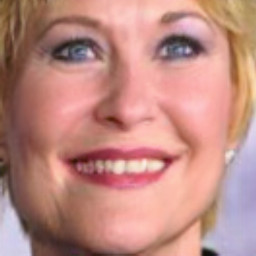} \hspace*{3pt}
  \framebox{\includegraphics[width=\boxwidth]{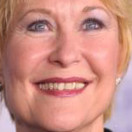}} \hspace*{3pt}
  \includegraphics[width=\boxwidth]{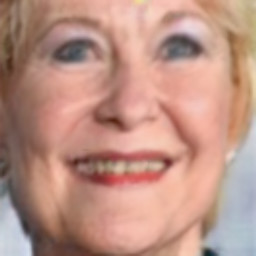} \\[1pt]
  \caption{Examples of transforming an approximately 55-year-old to older (65-year-old) and younger (towards 15-year-old). For each identity, row~1 shows the non-recursive transformation (applying the baseline transformer on each stage). Row~2 shows the partially-recursive transformation, with the double-trained \fromto{35}{25} transformer applied also to \fromto{45}{35} (the best chain, according to Alg.~\ref{alg:recu}). Row~3 shows the transformation with the most recursive steps, with the triple-trained \fromto{35}{25} transformer applied also to both \fromto{45}{35} and \fromto{55}{45} (the most compact chain, picked manually).}
  \label{fig:examples3}
\end{figure*}

In Fig.~\ref{fig:examples}, we show sample transformation paths, with a 15-year-old successively transformed to 65, and a 55-year-old transformed back to 15 and forward to 65. The results are visually at least at the level of \cite{antipov2017}. Additional results are shown in Figures \ref{fig:examples2} and \ref{fig:examples3}.

As most methods are concerned with singular transformations, our most relevant points of comparison are \cite{antipov2017} and \cite{lample2017}. However, both show results where several successive stages of development show effectively no changes at all, and no age measurements. Our results show greater variation between each stage, with not too many artifacts even at relatively low resolution. The improvements are presumably due to both the differences between reversible GANs and conditional GANs, and to the use of the transformer chain. The chain, of course, comes with additional computational cost.

In average, our compressed chain produced the target age in the validation set with an error less than $4.5\%$ (maximum of $3.2$~years) and standard deviations between $2.4$--$5.6$ in different age groups. The performance drop from the baseline due to compression was negligible. \cite{antipov2017} reports an estimation accuracy $17\%$ lower for synthetic images than natural ones. This may reflect the improvements in visual accuracy and variation in our model.

\section{Discussion and Conclusion}
\label{sec:conclusion}

In this paper, we showed that a transformer chain composed of reversible GAN image transformer modules can learn a complex multi-stage face transformation task. The domain-agnostic base algorithm is expected to generalize to other kinds of temporal progression problems. Notably, we found that a single transformer can carry out the face transformations \fromto{25}{35} and \fromto{35}{45} (and, with minor loss of accuracy, also \fromto{45}{55}). One might suspect this to be because not many visible changes happen during this time. However, the auxiliary age estimator can still discern the ages 25, 35, and 45 easily. As the estimator and transformer chains are independent, this indicates the presence of real changes that the networks capture in a systematic way, even when they are hard for humans to discern.

On some problem domains, the full chain may be relatively long and slow to train. Also, there is currently little control over the transformation paths, and the chain may not yield a range of varied outcomes in contexts where the developmental paths could diverge in different directions. Recent methods for improving the resolution of GAN-generated images \cite{karras2017} could be combined with our method. Follow-up work should evaluate the extent to which the layers of separate transformer modules can be shared, so as to reduce the total training time. More comprehensive evaluation of the models in both directions would likely find more compact chains. In semi-supervised setting, using a small number of paired examples would likely improve our results.

The code for replicating the results is available online: \url{https://github.com/AaltoVision/img-transformer-chain}.

\section*{Acknowledgments}

This research was supported by GenMind Ltd and the Academy of Finland grants 308640, 277685, and 295081. We acknowledge the computational resources provided by the Aalto Science-IT project.

\bibliographystyle{splncs03}
\bibliography{refs}

\end{document}